\documentclass{article}

\usepackage[utf8]{inputenc} 
\usepackage[T1]{fontenc}    
\usepackage{hyperref}       
\usepackage{url}            
\usepackage{booktabs}       
\usepackage{amsfonts}       
\usepackage{nicefrac}       
\usepackage{microtype}      
\usepackage{color}
\usepackage{graphicx}
\usepackage{subcaption}
\usepackage{xspace}
\usepackage{enumitem}
\usepackage{amsmath}
\usepackage{amssymb}
\usepackage{bm}
\usepackage{algorithm}
\usepackage{algorithmic}

\providecommand{\abs}[1]{\lvert#1\rvert}

\usepackage[accepted]{sysml2019}

\sysmltitlerunning{Exploiting Reuse in Pipeline-Aware Hyperparameter Tuning}

\begin{document}

\twocolumn[
\sysmltitle{Exploiting Reuse in Pipeline-Aware Hyperparameter Tuning}

\begin{sysmlauthorlist}
\sysmlauthor{Liam Li}{cmu}
\sysmlauthor{Evan Sparks}{determined}
\sysmlauthor{Kevin Jamieson}{uwash}
\sysmlauthor{Ameet Talwalkar}{cmu,determined}
\end{sysmlauthorlist}

\sysmlaffiliation{cmu}{Carnegie Mellon University}
\sysmlaffiliation{uwash}{University of Washington}
\sysmlaffiliation{determined}{Determined AI}

\sysmlcorrespondingauthor{Liam Li}{me@liamcli.com}

\sysmlkeywords{Machine Learning, SysML}

\vskip 0.3in

\begin{abstract}
Hyperparameter tuning of multi-stage pipelines introduces a significant computational burden.
Motivated by the observation that work can be reused across pipelines if the intermediate computations are the same, we propose a pipeline-aware approach to hyperparameter tuning.  Our approach optimizes both the \emph{design} and \emph{execution} of pipelines to maximize reuse.  We  design pipelines amenable for reuse by (i) introducing a novel hybrid hyperparameter tuning method called
gridded random search, and (ii) reducing the average training time in pipelines by adapting early-stopping hyperparameter tuning approaches.  We then realize the potential for reuse during execution by introducing a novel caching problem for ML workloads which we pose as a mixed integer linear program (ILP), and subsequently evaluating various caching heuristics relative to the optimal solution of the ILP.
We conduct experiments on simulated and real-world machine learning pipelines to show that a pipeline-aware approach to hyperparameter tuning can offer over an order-of-magnitude speedup over independently evaluating pipeline configurations.

\end{abstract}
]

\printAffiliationsAndNotice{\sysmlEqualContribution} 

\section{Introduction}
\label{sec:intro}
Modern machine learning workflows combine multiple stages of data-preprocessing, feature extraction, and supervised and unsupervised learning
~\citep{sanchez2013image,Feurer2015}.  The methods in each of these stages typically have configuration parameters, or \emph{hyperparameters}, that influence their output and ultimately predictive accuracy.  Although tools have been designed to speed up the development and execution of such complex pipelines~\citep{mllib,scikit-learn,KeystoneML}, tuning hyperparameters at various pipeline stages remains a computationally burdensome task. 

\begin{figure}[h]
\centering
\setlength\tabcolsep{.5pt}
\begin{tabular}{cc}

  \includegraphics[width=.23\textwidth, trim=1cm 0 1cm 0]{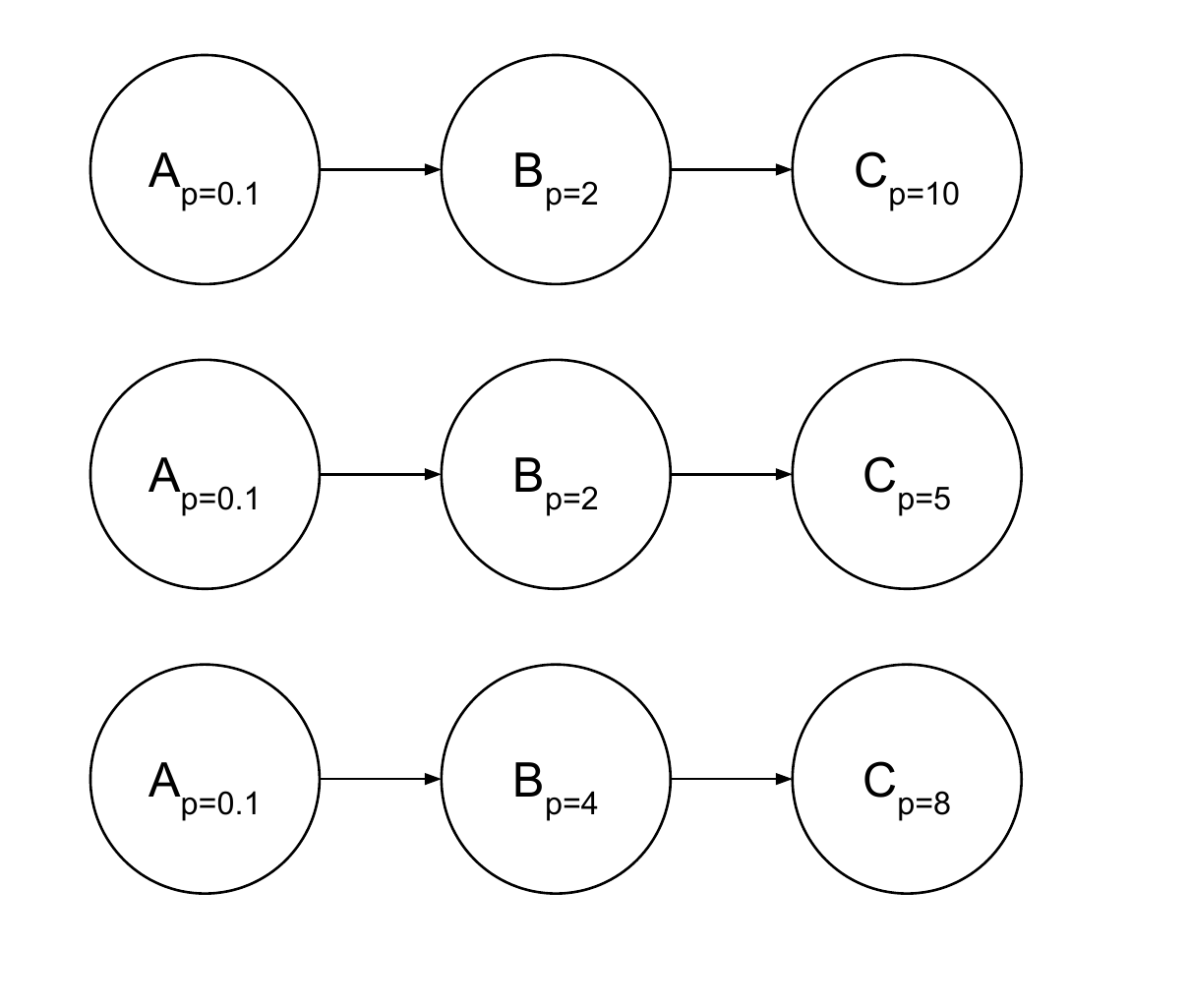} &
  \includegraphics[width=.23\textwidth, trim=1cm 0 1cm 0]{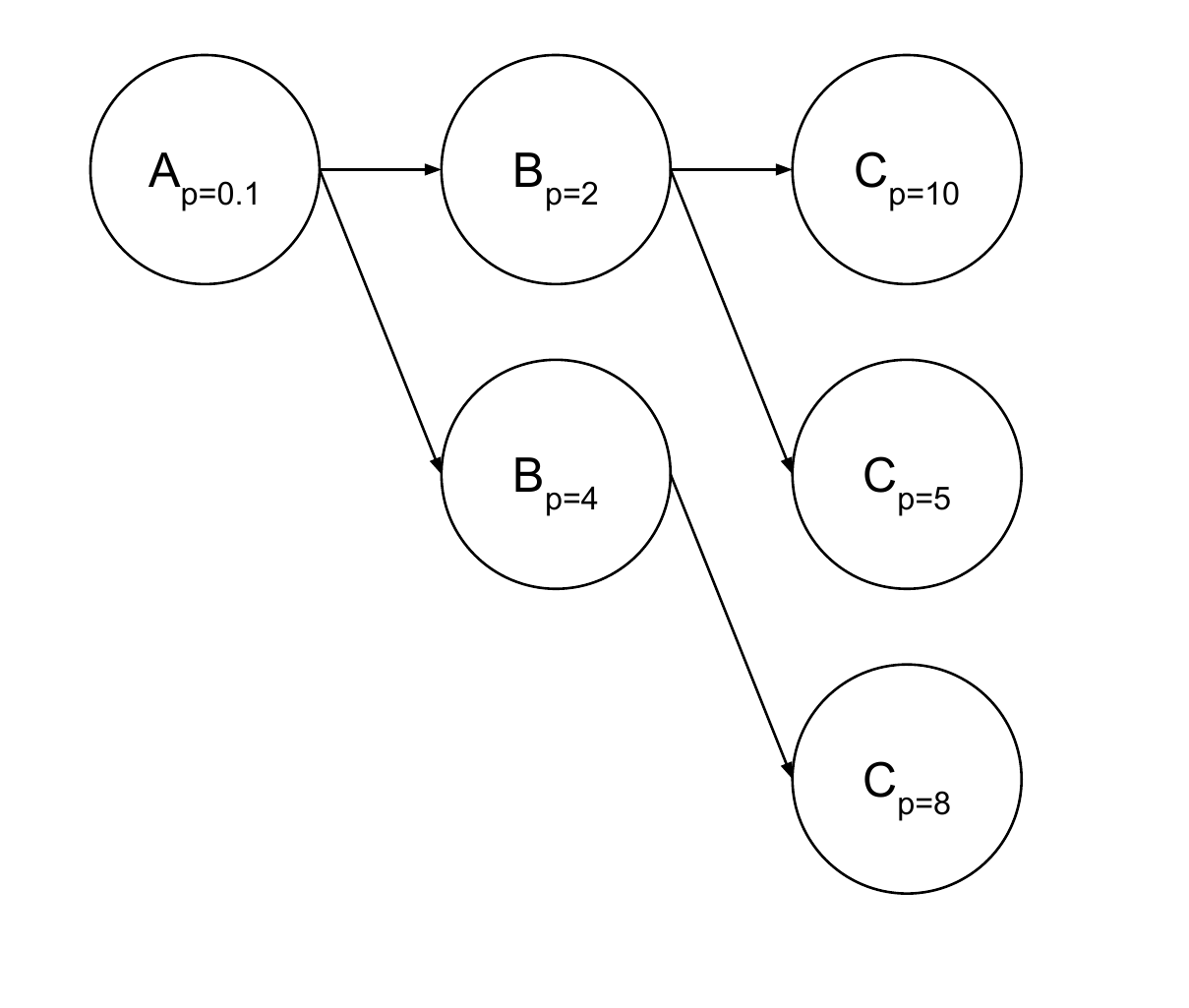} \\
  (a) & (b) \\
  \end{tabular}
\caption{Three pipelines with overlapping hyperparameter configurations for operations $A, B,$ and $C$. (a) Three independent pipelines. (b) Merged DAG after eliminating shared prefixes.}
\label{fig:combinedpipes}
\end{figure}

\begin{figure*}[th!]
\centering
\begin{subfigure}{0.9\textwidth}
	\centering
	\includegraphics[width=\textwidth]{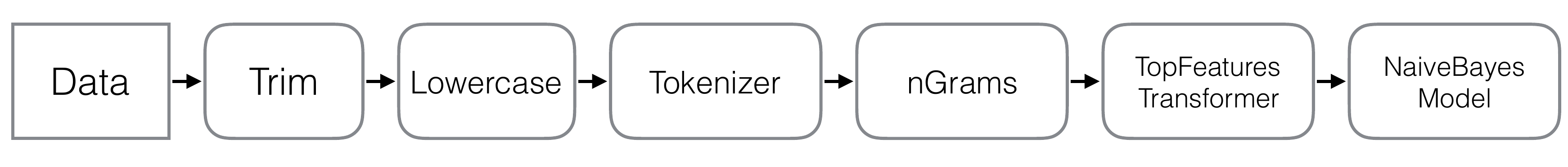}
	\caption{Example pipeline for text classification.}
		\vspace{.3cm}
	\label{fig:textpipeline}
\end{subfigure}
\begin{subfigure}{0.47\textwidth}
  \centering
  \includegraphics[width=\textwidth]{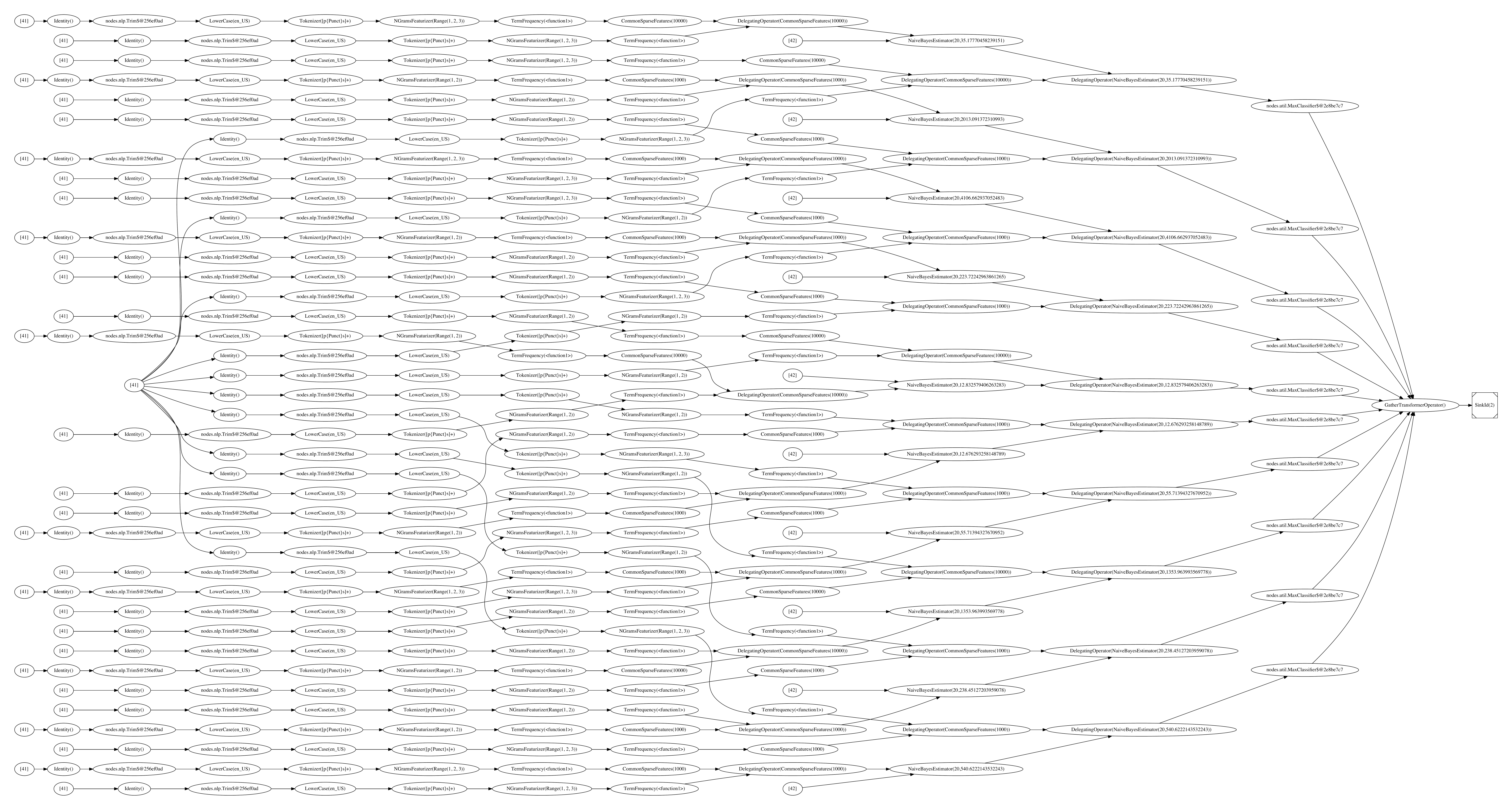}
  \caption{Pre-elimination pipeline.}
  \label{fig:pipesbeforeopt}
\end{subfigure}
\begin{subfigure}{0.47\textwidth}
  \centering
  \includegraphics[width=\textwidth]{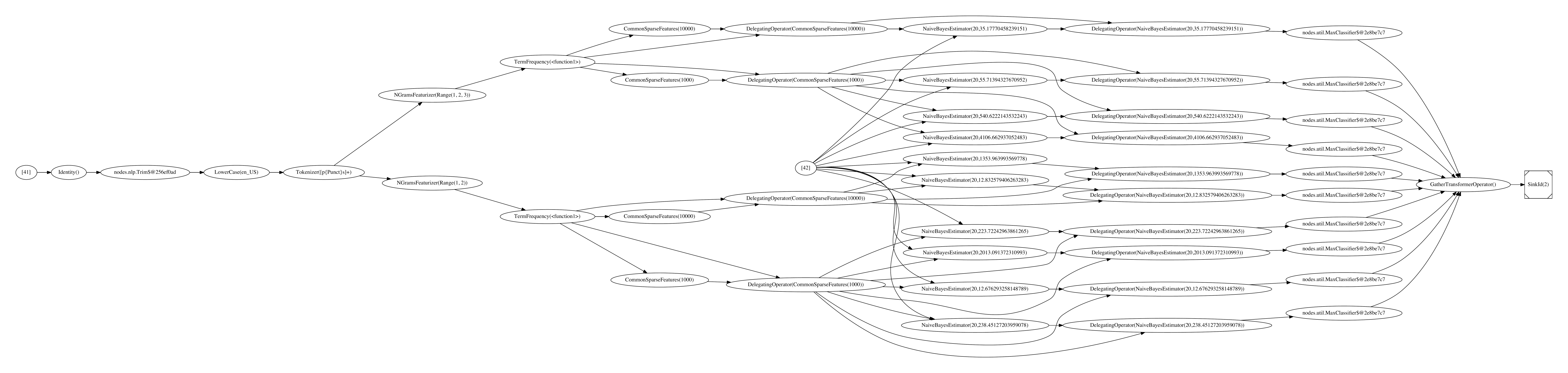}
  \caption{Post-elimination pipeline.}
  \label{fig:pipesafteropt}
\end{subfigure}
\caption{Ten sampled hyperparameter configurations for the pipeline in \ref{fig:textpipeline} before and after common prefix elimination.}
\label{fig:beforeandafter}
\end{figure*}

Some tuning methods are sequential in nature and recommend hyperparameter configurations one at a time for evaluation \citep[e.g.\ ][]{hutter2011, bergstra2011, Snoek2012}, while other  
batch mode algorithms \citep[e.g.\ ][]{hyperband, krueger2015fast, sabharwal2016selecting} evaluate multiple  configurations simultaneously.  In either case, holistically analyzing the functional structure of the resulting collection of evaluated configurations introduces opportunities to exploit computational reuse in the traditional black-box optimization problem.  
 
Given a batch of candidate pipeline configurations, it is natural to model the dependencies between pipelines via a directed acyclic graph (DAG).  We can eliminate redundant computation in the original DAG by collapsing shared prefixes so that pipeline configurations that depend on the same intermediate computation will share parents, resulting in a \emph{merged DAG}. We define \emph{pipeline-aware hyperparameter tuning} as the task of leveraging this merged DAG to evaluate hyperparameter configurations, with the goal being to exploit reuse for improved efficiency.  

Consider the example of the three toy pipelines in Figure~\ref{fig:combinedpipes}(a).
All of them have the identical configuration for operator $A$, and the first two have the same configuration for operator $B$, leading to the merged DAG illustrated in   
Figure~\ref{fig:combinedpipes}(b). 
Whereas treating each configuration independently requires computing node $A_{p=0.1}$ three times and node $B_{p=2}$ twice, operating on the merged DAG allows us to 
compute each node once.
Figure~\ref{fig:beforeandafter} 
 illustrates the potential for reuse via merged DAGs in a more realistic problem. 

In this work, we tackle the problem of pipeline-aware hyperparameter tuning, by optimizing both the
\emph{design} and \emph{evaluation} of merged DAGs to maximally exploit reuse. 
In particular, we identify the following
three core challenges, and our associated contributions addressing them.

\textbf{Designing Reusable DAGs.} Hyperparameter configurations are typically randomly sampled, often from continuous search spaces. The resulting merged DAGs consequently are extremely limited in terms of their shared prefixes, which fundamentally restricts the amount of possible reuse.  To overcome this limitation, we introduce a novel configuration selection method that encourages shared prefixes. In Section~\ref{ssec:gridded_random} we introduce our hybrid approach, called \emph{gridded random search}, which limits the branching factor of continuous hyperparameters, while performing more exploration than standard grid search. We show empirically that gridded random search performs comparably with random search, in spite of the known empirical limitations of grid search.

\textbf{Designing Balanced DAGs.} Modern  learning models are computationally expensive to train.  Therefore, the leaf nodes of the DAG can be more expensive to evaluate than intermediate nodes, limiting the impact of reuse.  To balance backloaded computation, 
we propose using hyperparameter tuning strategies that exploit early-stopping and partial training.  As we discuss in Section~\ref{ssec:warmstart}, these approaches can be interpreted as converting a single, expensive training node into a sequence of cheaper training nodes, resulting in a more balanced DAG.  In particular, we show how the Successive Halving algorithm~\citep{JamiesonTalwalkar2015} can be used to significantly reduce total training time.

\textbf{Exploiting Reuse via Caching.} Exploiting reuse requires an efficient cache strategy suitable for machine learning pipelines, which can have highly variable computational profiles at different nodes.  Although the general problem of caching is well studied in computer science \citep{sleater,McGeoch1991}, there is limited work on caching machine learning workloads, where the cost of computing an item can depend on the state of the cache.   We thus introduce the \emph{cache-dependent generalized paging problem} for this setting.  We show that the optimal cache strategy is the solution to a mixed integer linear program (Section~\ref{sec:caching}), and then empirically compare this optimal solution with faster heuristics.  Surprisingly, our results 
demonstrate that the least-recently used (LRU) cache eviction heuristic used in many frameworks (e.g., \texttt{scikit-learn} and \texttt{MLlib}) is not well suited for this setting, and we thus propose using WRECIPROCAL \citep{gunda2010nectar}, a caching strategy that accounts for both the size and the computational cost of an operation.
 
Finally, in Section~\ref{sec:evaluation}, we apply these three complementary approaches to real pipelines and show that pipeline-aware hyperparameter tuning can offer more than an \emph{order-of-magnitude} speedup compared with the standard practice of evaluating each pipeline independently.


\section{Related Work}
\label{sec:related_hp}
Hyperparameter tuning and caching are both areas with extensive prior work.  We highlight pertinent results in both areas in the context of reuse.  

\textbf{Hyperparameter Tuning Methods.}  Grid search is a classical approach for hyperparameter tuning, and it is naturally amenable to reuse via prefix sharing.  However, another classical method, random search, has been shown to be more efficient empirically than grid search \citep{Bergstra:2012ux}.  
 In Section~\ref{ssec:gridded_random}, we proposed a novel hybrid method designed to to combine the best of both of these classical methods (reuse from grid search and improved accuracy from random search). 

Several methods, e.g.,~\citet{JamiesonTalwalkar2015,hyperband,sabharwal2016selecting,pbtraining,gijsbers2018layered}, have exploited early-stopping to speed up hyperparameter search. 
These methods often reduce average training time per configuration by over an order-of-magnitude.  In Section~\ref{ssec:warmstart}, we aim to leverage this observation by deploying these early stopping methods 
in a pipeline-aware context to generate balanced DAGs amenable for reuse.  In our empirical results in Section~\ref{ssec:sha}, we focus on Successive Halving~\citep{karnin2013almost,JamiesonTalwalkar2015}, a simple early-stopping approach that is widely applicable, theoretically principled, and which has been shown to achieve state-of-the-art results on a variety of hyperparameter tuning tasks. 

\textbf{Caching with Perfect Information.}  One straightforward way of reusing intermediate computation is to save results to disk.  However,  the associated I/O costs are large, 
thus motivating  our exploration of efficient caching in memory 
as the primary method of realizing reuse in pipeline-aware hyperparameter tuning.  

Finding an optimal cache management policy is known as the \emph{paging problem}~\cite{bansal2012primal}. 
The classic paging problem assumes that all pages are the same size (i.e.\ memory requirement), and that all reads have the same cost (i.e.\ training time). Under these two assumptions, Belady's algorithm~\cite{belady1966study} is an optimal cache management policy for the perfect information setting, where the size and cost of all pages are known. 
If we relax the second assumption and allow for variable read costs, we now have the \emph{weighted paging problem}.  Lastly, if we further relax the first contraint to allow for variable page sizes, we have the \emph{generalized paging problem}, which encompasses the weighted paging problem.  The optimal perfect information policy for the generalized paging problem can be obtained by solving an integer linear program \cite{albers1999page}. 

The aforementioned problem settings assume all reads have fixed cost. In contrast, we consider a setting central to caching for machine learning pipelines, where the cost of computing an item depends on the state of the cache, i.e., it is cheaper to compute a node if any of the intermediate computation it depends on are in cache.  We study this  \emph{cache-dependent generalized paging problem} in Section~\ref{sec:caching}. 

\textbf{Online Caching.} In the previously described perfect information setting, the caching algorithm has access to 
the DAG, as well as the computational costs and memory
requirements of all nodes.  In contrast, in the online setting, the caching algorithm must make decisions based solely on
its prior experience.  Online algorithms often have significantly smaller computational overheads.  In Section~\ref{sec:caching}, motivated by computational concerns, we compare the optimal strategy to the cache-dependent generalized paging problem to  three online strategies:
(1) LRU: Least recently used items are evicted first. LRU is \emph{k-competitive} with Belady's algorithm \citep{sleater} for the classic paging problem with uniform size and cost.\footnote{An algorithm is $k$-competitive if its worst case performance is no more than a factor of $k$ worse than the optimal algorithm, where $k$ denotes the size of the cache.} 
(2) RECIPROCAL: Items are evicted with probability inversely proportional to the cost.  RECIPROCAL is $k$-competitive with the optimal caching strategy for the weighted paging problem \citep{randomalgo}.  
(3) WRECIPROCAL: Items are evicted with  probability inversely proportional to cost and directly proportional to size.  WRECIPROCAL is a weighted variant of RECIPROCAL that has been shown to work well for datacenter management \citep{gunda2010nectar}.  

For the classic paging problem, randomized online caching strategies outperform deterministic ones and are more robust \citep{Fiat1991}.  However, we include LRU in our comparison since it is widely used on account of its simplicity (including in some of the machine learning methods described below).  Additionally, there is a large body of work on efficient randomized online caching policies for the weighted paging problem and generalized paging problem \citep{bansal2008randomized}.  Studying the empirical effectiveness of these more recent and complex caching methods for tuning machine learning pipelines is a direction for future work.  

\textbf{Caching in Machine Learning.} There are existing methods that specifically address the issue of caching for machine learning pipelines. 
 Hyperparameter tuning algorithms like TPOT and FLASH offer LRU caching of intermediate computation.  However, as we show in Section~\ref{ssec:cachecomparison}, LRU is unsuitable for machine learning pipelines. \citet{scavenger2015} created a framework for users to specify the logic of which operations to cache for reuse, but did not provide an automated scheduler.  \citet{KeystoneML} addressed the issue of caching when evaluating a single pipeline, but did not account for the time-varying nature cache (i.e., did not allow for items to be evicted from cache to free up space).  Lastly, \citet{Xin2018Helix} and \citet{hagedorn2018} address reuse in an iterative workflow where a user may run similar pipelines multiple times in the course of development, but the exact nature of future workflows is unknown. 
In contrast, we address a setting where all the pipelines to be evaluated are known  and items can be stored or evicted from cache (Section~\ref{sec:caching}).

\section{DAG Formulation and Analysis}
\label{sec:dag}
We first formalize the DAG representation of the computation associated with a set of candidate pipeline configurations. Nodes within the DAG represent the computation associated with each stage of a single pipeline with specific hyperparameter settings and edges within the DAG represent the flow of the transformed data to another stage of the pipeline.  For a given node, the cost refers to the time needed to compute the output of the node, while the size refers to the memory required to store the output.  Finally, to construct the merged DAG, we merge two nodes if
(1) they share the same ancestors, indicating they operate on the same data; and
(2) they have the same hyperparameter settings, indicating they represent the same computation.  

We can use our DAG representation to gain intuition for 
 potential speedups from reuse.
For simplicity, we assume the cache is unbounded and we can store as many intermediate results as desired.  Additionally, 
 let $P$ be the set of all pipelines considered in the DAG.  Next, for a given pipeline $p$, let $V_p$ represent the nodes associated with each operator within
 the pipeline and $c(v)$ be the cost in execution time for a given node $v\in V_p$.   Then the total time required to evaluate 
the DAG without reuse is 
 $TP(P) = \sum_{p \in P} \sum_{v\in V_p} c(v).$

Let $P_{\text{merged}} = \cup_{p\in P} V_p$ represent the nodes in the merged DAG, where $\cup$ is the merge operator following the rules described above.  Then the runtime for the merged DAG is $TP(P_{\text{merged}}) = \smash{\sum_{v \in V_{P_\text{merged}}}} c(v)$, and the ratio, $\nicefrac{TP(P)}{TP(P_\text{merged})}$ is the speedup achieved from merging two pipelines.

For the purposes of illustration, if one holds $c(v)$ constant and assumes that each pipeline has the same length, $\abs{V}$, it is simple to see that $TP(P) = \abs{P}\abs{V}c(v)$. 
For the merged pipeline, if all the pipelines in the set are disjoint, then $TP(P) = TP(P_\text{merged})$ and the speedup is $1\times$, indicating no benefit from reuse. On the other hand, for the maximally redundant pipeline, where  all the pipelines differ only in the \emph{last} node,  the total execution time is $TP(P_\text{merged}) = (\abs{V} + \abs{P} - 1) c(v)$.  Hence, the maximum speedup ratio in this setting is  $\nicefrac{\abs{V}\abs{P}}{(\abs{V} + \abs{P} - 1)}$. In the limit, when all pipelines are long, speedups tend to $\abs{P}$, while if the number of trajectories dominate, speedups tend to $\abs{V}$.  

Next, if we relax the assumption that $c(v)$ is fixed, and assume instead that merged nodes are expensive relative to the \emph{last} stage of the pipeline, the speedups will approach $|P|$,  since a pipeline with more expensive intermediate nodes is similar to a longer pipeline with fixed cost.  Alternatively, if we assume that the last stage of the pipeline is expensive relative to intermediate stages, we can interpret this as a long pipeline where many stages are disjoint, and the speedups will approach $1\times$.  
Hence, to maximize speedups from reuse, we need to optimize the design of DAGs to promote prefix sharing and a balanced computational profile with limited time spent on training leaf nodes.  

\section{Designing `Good' DAGs}
\label{sec:promoting_reuse}
In this section, we focus on the two design challenges discussed in Section~\ref{sec:intro}:
(i) designing reusable DAGs with prefix sharing across pipelines when faced with search spaces with continuous hyperparameters, and (ii) designing balanced DAGs in spite of backloaded computation due to computationally expensive model training routines. 
\subsection{Reusable DAGs via Gridded Random Search}
\label{ssec:gridded_random}

\begin{figure}[h]
\centering
\includegraphics[width=0.45\textwidth]{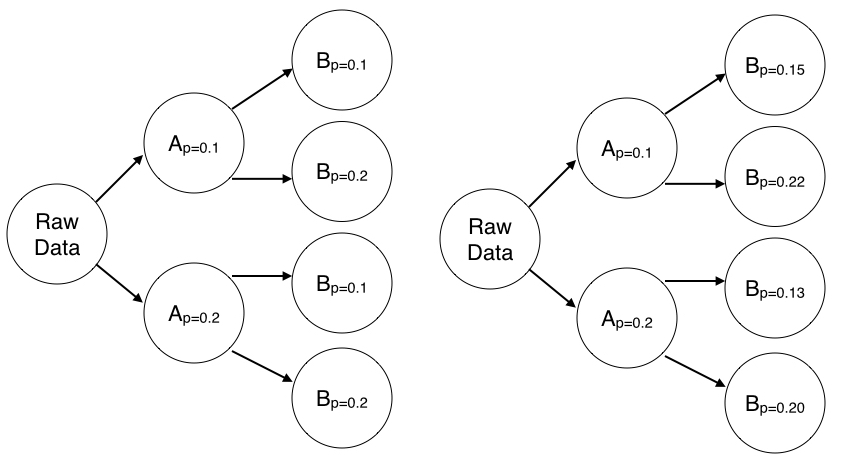}
\caption{Comparison of grid search versus gridded random search.  Grid search considers two different values for each hyperparameter dimension.  In contrast, gridded random considers two hyperparameter values for operator $A$ and four different values for operator $B$, while still maintaining a branching factor of  two for the DAG.  Note that random search would consider four different values for $A$ and four different values for $B$ but allow no potential for reuse.  }
\label{fig:gridded_random}
\end{figure}

As discussed in Section~\ref{sec:related_hp}, grid search discretizes each hyperparameter dimension, while random search samples randomly in each dimension.  Gridded random is a hybrid of the two approaches that encourages prefix sharing by controlling the branching factor from each node in a given stage of the pipeline, promoting a narrower tree-like structure.  We show a visual comparison between grid search and gridded random search in Figure~\ref{fig:gridded_random}.  The key difference from grid search is that the children of different parent nodes from the same level are not required to have the same hyperparameter settings.  In effect each node performs a separate instance of gridded random search for the remaining downstream hyperparameters.  

Gridded random search introduces a tradeoff between the amount of potential reuse and coverage of the search space.  Generally, the branching factor should be higher for nodes that have more hyperparameters in order to guarantee sufficient coverage.  We compare gridded random search to random search in Section~\ref{ssec:micro_grid} and show that they perform comparably, despite the former's limited coverage of the search space relative to the latter. 

\subsection{Balanced DAGs via Early-Stopping}
\label{ssec:warmstart}

As discussed in Section~\ref{sec:dag}, a DAG's potential for reuse is  severely limited by pipelines with long training times.
For backloaded pipelines, we can increase the potential for reuse by reducing the average training time per configuration, e.g., by employing iterative early-stopping hyperparameter tuning algorithms when evaluating a batch of configurations.  
\begin{algorithm}[h]
\begin{algorithmic}
\STATE\textbf{require:} define \texttt{step}, \texttt{prune}, \texttt{populate}
\STATE\textbf{input:} initial population $P$, generations $G$, 
\STATE
\FOR{$i=1,\dots,G$}
\STATE $\texttt{step}(P)$ \COMMENT{Partially train configs in P.}
\STATE $P \leftarrow \texttt{prune}(P)$ \COMMENT{Eliminate underperforming configurations.}
\STATE $P \leftarrow \texttt{populate}(P)$ \COMMENT{Warmstart child configurations from successful parents.}
\ENDFOR
\end{algorithmic}
\caption{Generic Early-Stopping Tuning Algorithm.}
\label{alg:warmstart}
\end{algorithm}

Algorithm~\ref{alg:warmstart} presents a generic approach which encompasses the early-stopping methods mentioned in Section~\ref{sec:related_hp}.  
Inspired by common paradigms seen in evolutionary strategies \citep{schwefel1993evolution,hansen1995adaptation,back2013evolution}, Algorithm~\ref{alg:warmstart} evolves $G$ generations from an initial population $P$ using user-defined functions \texttt{step}, \texttt{prune}, and \texttt{populate}.  $P$ is an initial set of pipeline configurations from a user specified search space.  The subroutine \texttt{step} takes the current population of pipelines with associated states (i.e.\ trained weights) and allocates training resources to each pipeline.  The \texttt{prune} subroutine evaluates the partially trained models using a user defined objective function and prunes the population based on observed scores.  Finally, the \texttt{populate} subroutine spawns the next generation of models to populate the remaining branches, with the option of warmstarting descendants with models from $P$.  

Algorithm~\ref{alg:warmstart} can be viewed as a subdivision of the training stage, which is usually just a single stage of a pipeline, into multiple partial training stages. In effect, it gives us a multi-stage pipeline with more balanced computational costs.  
We provide a concrete example in Section~\ref{ssec:sha}, where we present the Successive Halving algorithm \citep{karnin2013almost,JamiesonTalwalkar2015}  as an early-stopping algorithm and analyze its effect in producing balanced DAGs.

\section{Exploiting Reuse via Caching}
\label{sec:caching}
Given the discussion in Section~\ref{sec:promoting_reuse} on optimizing the design of DAGs, we now tackle the third challenge of realizing this potential through efficient caching of intermediate results.
Specifically, we formulate the cache-dependent generalized paging problem introduced in Section~\ref{sec:related_hp} as a mixed integer linear program (ILP).  The solution to this ILP provides the optimal cache policy in the perfect information setting.
\subsection{Setup}
We start with a few definitions. Let $G$ denote a DAG consisting of vertices $V$ and edges $E$.
The \emph{sources} of the graph are any $v_x \in V$ such that $v_x \ne v_j \; \forall (v_i,v_j) \in E$, while the \emph{sinks} are any $v_x \in V$ such that $v_x \ne v_i \; \forall (v_i,v_j) \in E$.

An \emph{execution plan} is a sequence of paths from sources to sinks that together include all edges in $E$.    The execution plan can be viewed as all of the outputs the program would have to compute in the absence of a cache, with the only item in working memory being an operator's results and its direct inputs.  As we progress through the execution plan, we call the path of the current node the \emph{active path}.
If we consider the graph in Figure~\ref{fig:combinedpipes}, then 
the execution plan matching the $A_{p=0.1}B_{p=2}C_{p=10}C_{p=5}B_{p=4}C_{p=8}$ depth-first traversal is:
\[
A_{p=0.1}B_{p=2}C_{p=10},A_{p=0.1}B_{p=2}C_{p=5},A_{p=0.1}B_{p=4}C_{p=8}\]

Let ($\bm{i}$) be a vector of length $T$ where $i_t$ is the $t$th node of the execution plan.  
At each time $t$, the cost of computing a node is given by the state of the cache. If there exists a node between $i_t$ and the ``active'' sink that is cached, then the time devoted to $i_t$ is 0, otherwise its $c_t = C(i_t)$, where $C$ is a cost vector mapping each node to a non-negative real number.  Intuitively, if a successor step of the current node is saved in cache, there is no need to perform the computation at the current node.   Similarly, let $m_t = M(i_t)$ be the memory required to cache $i_t$, where $M$ is a vector mapping each node to a non-negative real number.  

Let $\Delta$ be a $V\times T$ matrix representing a sequence of valid actions on the cache, and define $\Delta$ as follows
\[
	\Delta_{j,t} = 
	\begin{cases}
		1 & \text{if} \ j \ \text{is \emph{added} to the cache at time t}\\
		-1 & \text{if} \ j \ \text{is \emph{removed} from the cache at time t}\\
		0 & \text{otherwise}
	\end{cases}
\]

Additionally, let $X\in\{0,1\}^{V\times T}$ represent the state of the cache through time with entries 
\[
    X_{j,t} = \sum_{s < t} \Delta_{j,s} \,\forall j \in V
\]

Then, the set of \emph{valid} actions on the cache is severely limited; $\Delta_{j,t}$ may only be positive if $j = i_t $, and may only be negative if $X_{j,t-1} > 0$. That is, the system can only add the active item, and can only evict items that are currently in the cache. 

Let $A \in \{0,1\}^{V \times T}$ indicate whether a downstream operator influences the state of the cache. That is,

\[
    A_{j,t} = 
    \begin{cases}
        1 &\text{if} \ j \ \text{is on the active path between}\\
        & \ i_t \ \text{and the sink}\\
	    0 &\text{otherwise}
	\end{cases}
\]
Specifically, $A_{j,t}=1$ for the active node and all of its successors along the active path.

\subsection{Cache-Dependent Generalized Paging Problem}

We now formally define our caching problem of interest:

\textbf{Definition 1. (Cache-Dependent Generalized Paging)} \emph{The goal is to find the cache policy, $\Delta$, that minimizes the following expression:}
\begin{equation*}
    \begin{aligned}
        \underset{X}{\text{minimize}} & & \sum_{t \in T} c_t \text{max}(0, 1 - X_t^\top A_t)\\
        \text{subject to} && \Delta_{j,0} = 0 & \quad \forall j \in V,\\
        && \Delta_{j,t} \leq 0 & \quad \forall j \neq i_t,\\
        && \Delta_{i_t,t} \geq 0, \\
        && -1 \le \Delta_{j,t} \le 1 & \quad \forall j,t, \\
        && \Delta_{j,t} \in \mathbb{Z} & \quad \forall j,t, \\
        && X_{j,t} = \sum_{k=1}^t \Delta_{j,k} & \quad \forall j, \\
        && 0 \leq X_{j,t} \leq 1 & \quad \forall j,t, \\
        && 0 \leq \sum_{i=1}^n X_{i,t} m_i \leq M, & \quad \forall t 
    \end{aligned}
\end{equation*}

That is, the goal is to minimize the total runtime of the DAG where the runtime is dependent on the structure of the graph ($A_t$) and the state of the cache at each point in time ($X_t$). The cost of evaluating a node is included in the runtime if it is not in cache and must be evaluated to reach the sink ($\text{max}(0, 1- X_t^\top A_t) > 0$). The first four constraints on $\Delta$ correspond to the following requirements: (1) the cache must be initially empty; (2) only the active node may be added to cache; (3) only non-active nodes may be removed from the cache; and (4) $\Delta$ is bounded above and below by $1$ and $-1$. The fifth constraint requires $\Delta$ to take on integral values, making this a mixed-integer linear program, as opposed to a constrained linear program. Next, the equality constraint simply says that $X$ is the cumulative sum of the \emph{changes} to the cache up to time $t$. The final constraint requires that at any time $t$, the total size of the objects in cache must be below a positive real number, $M$. 

The objective function is convex in $X$. This can be seen as follows: $0$ is convex, $1 - X_t^\top A_t$ is convex, and the $max$ of any two convex functions is convex, as is the same function scaled by a non-negative number. Finally, the sum of convex functions is also convex. Furthermore, each of the constraints on the program are either linear or equality constraints, with the exception of one integral constraint. Hence, this program is a mixed integer linear program (ILP). 

Note that extending this formulation to account for a scenario where evicted items can be saved to disk and reloaded later is fairly simple.  First, we should save a result to disk if reading from disk is cheaper than recomputing from the root node.  If it is cheaper to read from disk, we modify the cost $c_t$ of computing a node after the first time it is seen to be the cost of reading from disk.  

It is well known that solving ILPs is NP-hard \citep{papadimitriou1998combinatorial}.  
Nonetheless, in Section~\ref{ssec:ilpscaling}, we investigate when it is feasible to solve the ILP to recover the optimal policy.  Then, equipped with this information, we use the optimal cache policy to inform the selection of an online cache strategy with smaller overhead when executed on larger DAGs.

\section{Analysis of Individual Proposed Solutions}
In this section, we delve into our proposed solutions to each of the three core challenges limiting reuse in machine learning pipelines.  Specifically, in Section~\ref{ssec:micro_grid} we compare gridded random search with random search on 20 OpenML \citep{OpenML2013} datasets; we present a concrete example of using an early-stopping method to reduce average training time in Section~\ref{ssec:sha}; and evaluate our optimal caching strategy on simulated DAGs in Section~\ref{ssec:cache_eval}.

\subsection{Gridded Random Search versus Random Search}
\label{ssec:micro_grid}

 We evaluate the performance of gridded random search compared to standard random search on 20 OpenML classification datasets.   
The search space we consider includes the following pipeline stages: (1) Preprocessor: choice of none, standardize, normalize, or min/max scaler; 
(2) Featurizer: choice of PCA, select features with top percentile variance, or ICA; 
(3) Classifier: choice of linear classifier using stochastic gradient descent, random forests, or logistic regression.  
For each dataset, all four preprocessors are considered, but a single featurizer and a single classifier are randomly selected.  
For gridded random search, the branching factor in the first stage is 4, one for each preprocessor type; the branching factor is 5 per node for the selected featurizer; and then 5 per featurizer node for the selected classifier. This results in 100 total pipelines for gridded random search.  Appendix~\ref{appendix:grid} provides more details on  the selection process for the OpenML datasets and the search spaces used for each component.

\begin{table}[h!]
\begin{center}
\tiny
\resizebox{\linewidth}{!}{
\begin{tabular}{lll|rrr}
\hline
Dataset ID & Featurizer & Classifier & Mean & 20\% & 80\% \\
\hline
OpenML 182 & SelectPercentile & LogisticReg & 0.1\% & 0.0\% & 0.0\% \\
OpenML 300 & SelectPercentile & SVC & 0.6\% & 0.6\% & 0.6\% \\
OpenML 554 & PCA & LogisticReg & 0.0\% & 0.1\% & 0.0\% \\
OpenML 722 & PCA & RandForest & -0.1\% & -0.1\% & -0.1\% \\
OpenML 734 & SelectPercentile & LogisticReg & 0.0\% & 0.1\% & 0.0\% \\
OpenML 752 & SelectPercentile & LogisticReg & -0.0\% & 0.0\% & -0.0\% \\
OpenML 761 & SelectPercentile & LogisticReg & 0.6\% & 2.1\% & 0.0\% \\
OpenML 833 & SelectPercentile & RandForest & -0.0\% & 0.1\% & -0.1\% \\
OpenML 1116 & PCA & RandForest & -0.2\% & -0.2\% & -0.3\% \\
OpenML 1475 & SelectPercentile & SVC & -0.8\% & -0.9\% & -0.9\% \\
OpenML 1476 & PCA & SVC & 0.3\% & -0.4\% & 0.0\% \\
OpenML 1477 & PCA & SVC & 0.4\% & -0.8\% & 0.1\% \\
OpenML 1486 & PCA & SVC & 0.2\% & 0.3\% & 0.3\% \\
OpenML 1496 & FastICA & LogisticReg & 0.0\% & 0.1\% & -0.0\% \\
OpenML 1497 & PCA & RandForest & -0.3\% & -0.1\% & 0.2\% \\
OpenML 1507 & SelectPercentile & SVC & 0.1\% & 0.2\% & 0.0\% \\
OpenML 4538 & SelectPercentile & RandForest & -0.0\% & -0.4\% & 0.3\% \\
OpenML 23517 & SelectPercentile & RandForest & -0.0\% & 0.0\% & -0.1\% \\
OpenML 40499 & PCA & LogisticReg & 2.9\% & 6.1\% & 1.5\% \\
OpenML 40996 & SelectPercentile & RandForest & -0.0\% & -0.1\% & -0.1\% \\
\end{tabular}
}
\end{center}
\caption{For each OpenML dataset, we randomly sample a featurizer and classifier to construct the search space.  The right three columns show the difference in aggregate statistics between random search and gridded random search.  Positive values indicates the performance of random is better than that of gridded random. For each task, statistics for the best hyperparameter setting found over 100 pipelines are computed over 10 independent runs.}
\label{tab:gridvsrandom}

\end{table}

For each dataset, we evaluated a set of 100 pipelines sampled using either random or gridded random and recorded the best configuration for each sampling method.  To reduce the effect of randomness, we averaged results across 10 trials of 100 pipelines for each dataset.  
Our results in Table~\ref{tab:gridvsrandom} show that on average gridded random search converges to similar objective values as standard random search on a variety of datasets.
With the exception of one task, the differences in performance between random and gridded random are all below 1\% accuracy.   While by no means exhaustive, these results demonstrate the viability of using gridded random search to increase reuse.  

\subsection{Balancing DAGs via Successive Halving}
\label{ssec:sha}

\begin{figure*}[t]
\centering
\includegraphics[width=0.7\textwidth]{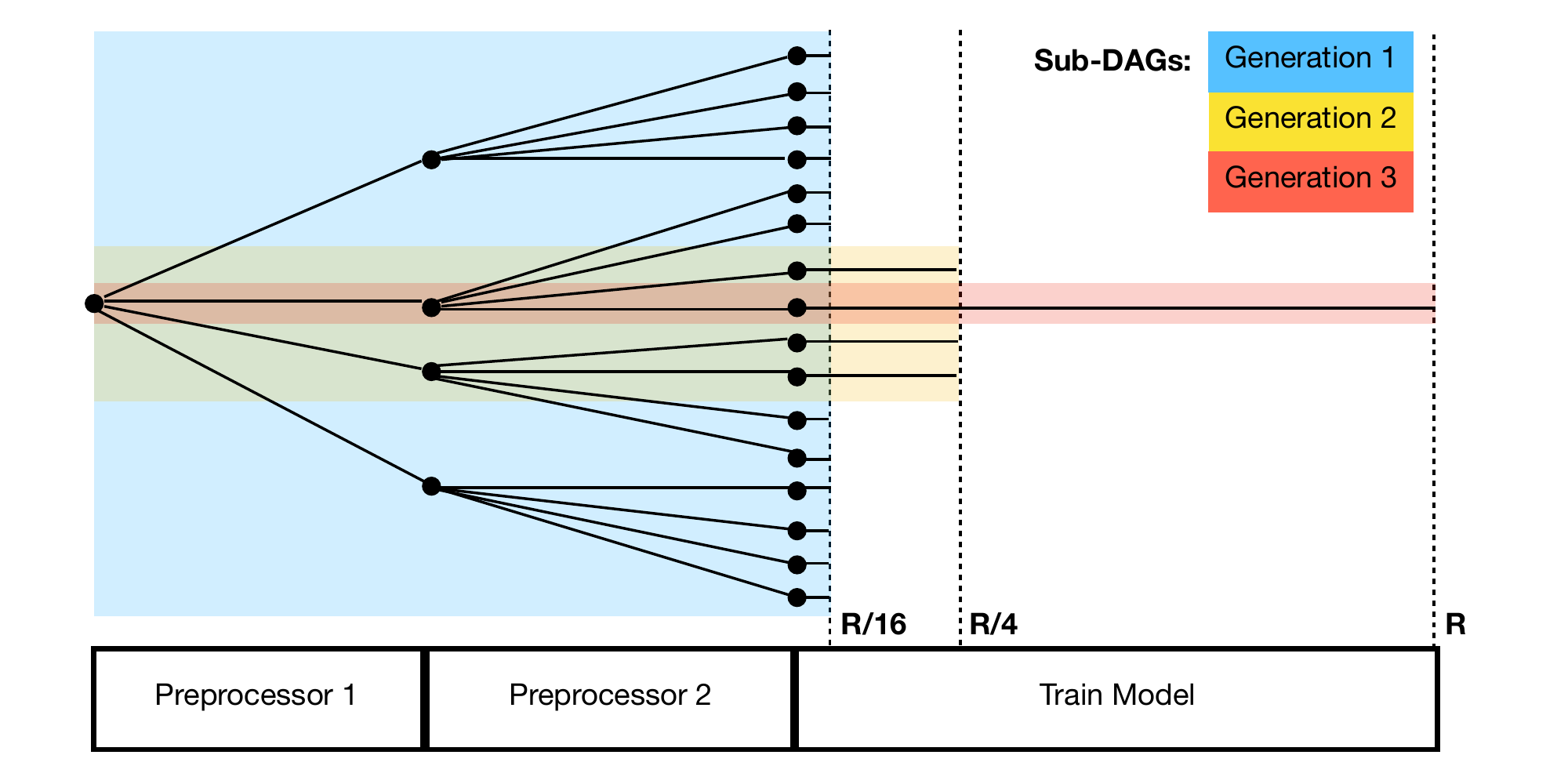}
\caption{In this DAG, all pipelines start with the raw data in the root node and proceed to subsequent steps of the pipeline with a branching factor of 4 for a total of 16 pipelines.   The lengths of the edges correspond to the time needed to compute the subsequent node.  Hence, for the depicted pipelines, the time needed to train a model on $R$ resource is equal to the time needed to compute the two preprocessing steps.  Therefore, without early-stopping, training time accounts for over half of the total time needed to execute the entire DAG.  Switching over to Successive Halving (SH) with elimination rate $\eta=4$ and generations $G=3$, the shaded blocks indicate the resulting pruned DAGs after each generation.   The lengths of the edges in the `Train Model' phase correspond to resources allocated to those configurations by SH: 16 configurations are trained for $R/16$ in the first generation, 4 configurations for $R/4$ in the second generation, and one configuration for $R$ in the last generation.  Hence, the total training time required for SH 
is $3R$, which is over $5\times$ smaller than the $16R$ needed without early-stopping, leading to a DAG with more frontloaded computation.  Note that   SH is normally run with larger DAGs (more pipelines) and more aggressive early-stopping rates, which, as we show in Section~\ref{ssec:timit}, further increase speedups from reuse.}
\label{fig:sha_dag}
\end{figure*}
\begin{figure*}[th!]
\centering
\begin{subfigure}{0.4\textwidth}
	\centering
	\includegraphics[width=\textwidth]{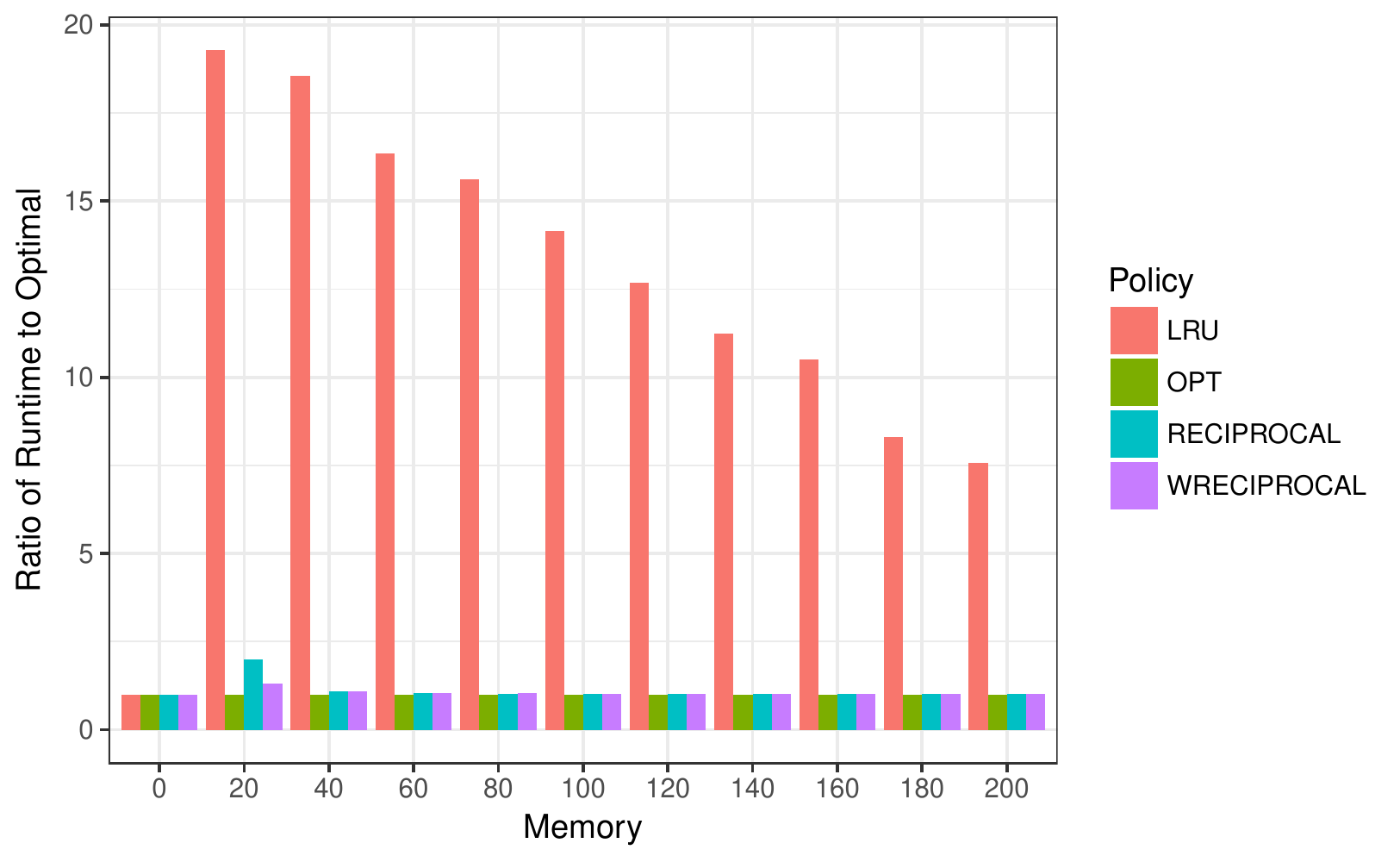}
	\captionof{figure}{All nodes have size 10; root node has cost 100 and all other nodes have cost 1.}
	\label{fig:eventreeload}
\end{subfigure}
\hspace{0.2cm}
\begin{subfigure}{0.4\textwidth}
	\centering
	\includegraphics[width=\textwidth]{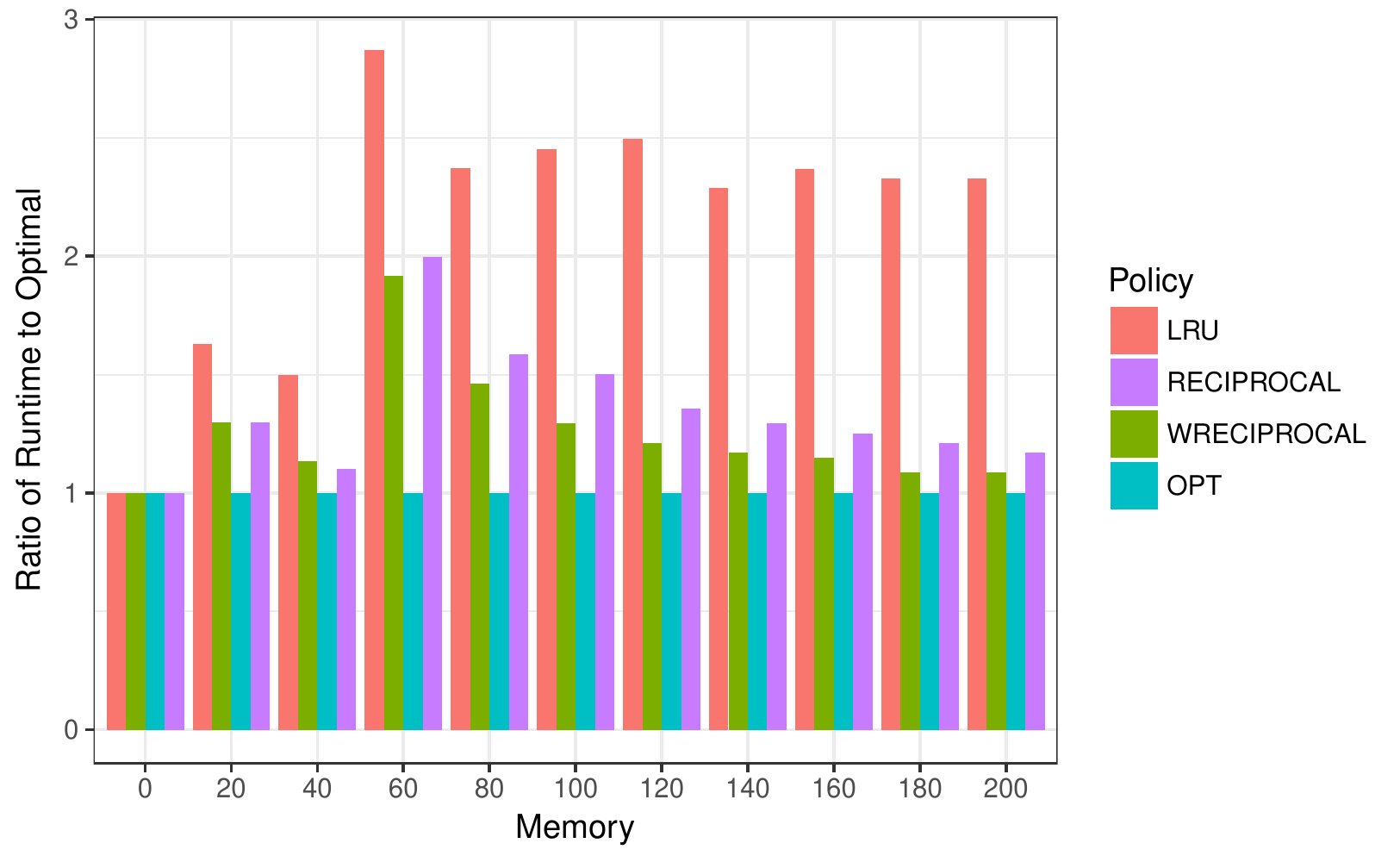}
	\captionof{figure}{For each node, size is randomized to be 10 or 50 and cost is randomized to be 1 or 100.}
	\label{fig:uneventreeload}
\end{subfigure}
\caption{A comparison of cache management policies on a 3-ary tree with depth 3. }
\end{figure*}
We now demonstrate the impact of early-stopping on generating balanced DAGs, focusing on Successive Halving (SH).
Algorithm~\ref{alg:sha} shows SH formulated as an instantiation of Algorithm~\ref{alg:warmstart}.
SH with an elimination rate of $\eta$ begins with a set of configurations $P$ and proceeds as follows: allocate a small amount of resource to each configuration, keep the top $1/\eta$ configurations, and repeat with the remaining configurations, while increasing the resource per configuration in each generation by a factor of $\eta$.  Notably, the specified inputs $\eta$, $G$ (generations) and $R$ (maximum resource per configuration) define the minimum resource per configuration is $r=R\eta^{-(G-1)}$. 
Moreover, when $G=1$, SH reduces to random search since the minimum resource is $r=R$.
\begin{algorithm}[h]
\begin{algorithmic}
\STATE \textbf{input:} $\enspace$ set of configurations $P$, maximum resource per configuration $R$, elimination rate $\eta$, generations $G \geq 1$
\STATE

\STATE \texttt{assert} $n\geq\eta^{G-1}$ so that at least one configuration will \\
\hspace{0.5cm}be trained for $R$ resource
\STATE $r = R\eta^{-(G-1)}$

\FOR{$g$ in $1,\dots G$}
\STATE $\texttt{step}$: Train each configuration in $P$ for  $r\eta^{g-1}$ resource
\STATE $P \leftarrow \texttt{prune}(P)$: evaluate and keep top $\lfloor|P|/\eta\rfloor$ \\
\hspace{1cm} configurations
\STATE $P \leftarrow \texttt{populate}(P)$: warmstart with remaining \\
\hspace{1cm} configurations
\ENDFOR
\end{algorithmic}
\caption{Successive Halving Algorithm.}
\label{alg:sha}
\end{algorithm}

As an example of how SH can be combined with pipeline-aware hyperparameter tuning, consider the DAG in Figure~\ref{fig:sha_dag} where the time needed for the preprocessing steps in the pipeline is the same as the training time. Assume we run SH with $|P|=16$, $\eta=4$, for $G=3$ generations.    SH begins with the full DAG containing 16 pipelines and traverses the DAG, while training leaf nodes (i.e., a classifier) with the initial minimum resource $r=R/16$ (blue sub-DAG).  In the next generation, it increases the resource per leaf node to $R/4$ and traverse the pruned DAG with 4 remaining configurations (yellow sub-DAG).  Finally, in the third generation a single pipeline remains, and it is trained for the maximum resource $R$ (red sub-DAG).    Hence in total, SH use $3R$ resources for training compared to the $16R$ that would be needed without early-stopping.   If training times scaled linearly, then total training time is reduced by over $5\times$ with SH.  Visually, this reduction in training time is evident in Figure~\ref{fig:sha_dag}, where the computational profile for the sub-DAGs is much more frontloaded. 

In general, for a maximum resource of $R$ per configuration, a total budget of $nR$ is needed to evaluate $n$ configurations without early-stopping.  In contrast, Successive Halving requires a budget equivalent to $nR\eta^{-(G-1)}G$, where $nR\eta^{-(G-1)}$ indicates the total resource allocated in each generation. Hence,  SH can be used to balance backloaded computation by reducing the total training time by a factor of $\eta^{G-1}/G$ relative to no early-stopping. Note that   SH is normally run with larger DAGs and more aggressive early-stopping rates, which as we show in Section~\ref{ssec:timit}, can drastically increase the speedups from reuse on a real world pipeline optimization task.  

\begin{figure*}[th!]
\centering
\begin{subfigure}{0.3\textwidth}
\includegraphics[width=1.1\textwidth]{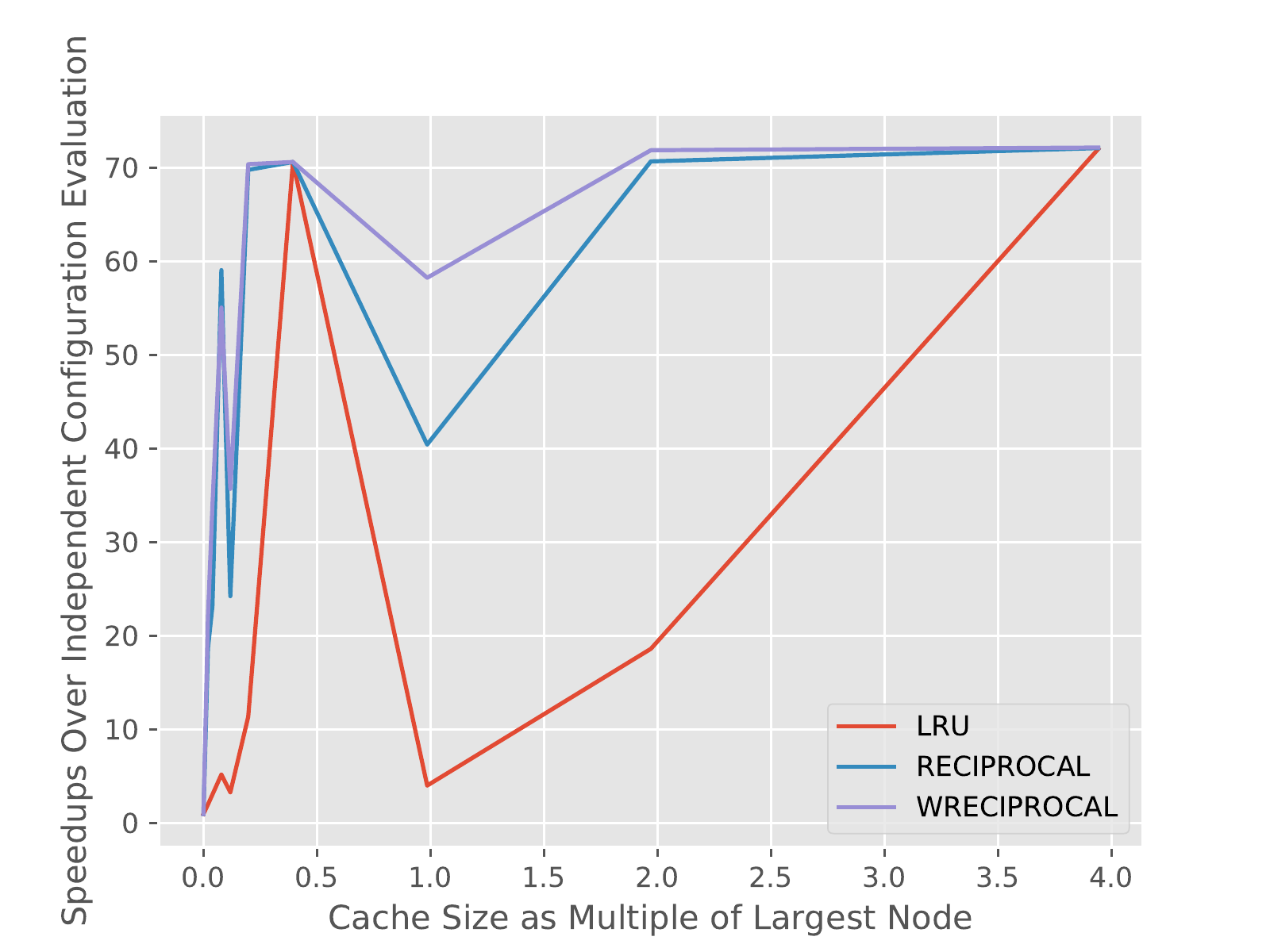}
\caption{20 Newsgroups}
\end{subfigure}
\hspace{1cm}
\begin{subfigure}{0.3\textwidth}
\includegraphics[width=1.1\textwidth]{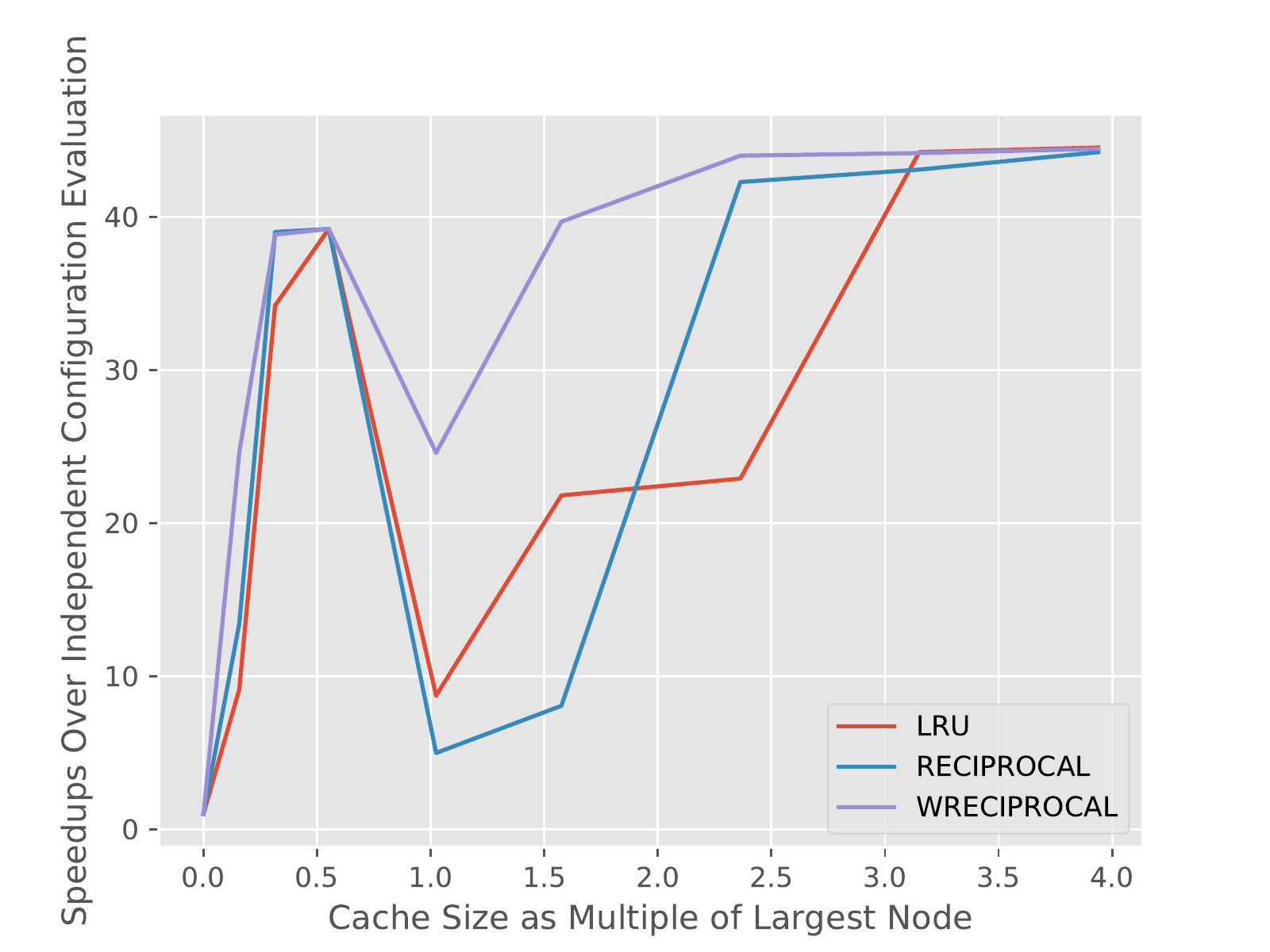}
\caption{Amazon Reviews}
\end{subfigure}
\caption{Effect of reuse on two real-world text processing hyperparameter tuning tasks.  Speedups over independent configuration evaluation when evaluating a DAG with 100 pipelines are shown for each cache policy given a particular cache size.   }
\label{fig:uniform}
\end{figure*}
\subsection{Evaluating Caching Strategies}
\label{ssec:cache_eval}
We start by quantifying the runtime required for solving the ILP (Definition~$1$) to arrive at the optimal cache policy for the perfect information setting.  We then use the optimal caching strategy to inform the selection of an efficient caching strategy with lower computational overhead. 

\subsubsection{Limitations on Solving the ILP}
\label{ssec:ilpscaling}
\begin{table}[h]
\centering
\tiny
\resizebox{0.9\linewidth}{!}{
\begin{tabular}{ccccccc}
	\hline \\[-1.8ex]
	\hline \\[-1.8ex]
	$k$ & $d$ & $p$ & $n$ & $T$ & Size of $X$ & Runtime (s)\\
	\hline \\
	2 & 2 & 4   & 7    & 12    & 84      & 0.5\\
	2 & 3 & 8   & 15   & 32    & 480     & 1.3\\
	2 & 4 & 16   & 31   & 80   & 2480     & 3.8\\
	2 & 5 & 32  & 63   & 192   & 12096    & 12.9\\
	3 & 2 & 9   & 13   & 27    & 351     & 3.5\\
	3 & 3 & 27   & 40   & 108   & 4320    & 70.9\\
	3 & 4 & 81  & 121  & 405   & 49005    & x\\
	3 & 5 & 243  & 364  & 1458  & 530712   & x\\
	4 & 2 & 16   & 21   & 48   & 1008     & 27.3\\
	4 & 3 & 64  & 85   & 256   & 21760    & x\\
	4 & 4 & 256  & 341  & 1280  & 436480   & x\\
	4 & 5 & 1024 & 1365 & 6144 & 8386560  & x\\
	\hline
\end{tabular}
}
\caption{Time required to solve the ILP for perfect $k$-ary trees of depth $d$.  For each $k$ and $d$, $p$ counts the number of total pipelines, $n$ the number of nodes, and $T$ the length of the execution plan.  As in (Definition~$1$), $X$ represents the state of the cache.  Observations with an `x' timed out after 5 minutes. }
\label{tab:slowtrees}
\end{table}

We evaluate the runtime required to solve the ILP on synthetic DAGs designed to look like common hyperparameter tuning pipelines in terms of size and computational cost.  We consider DAGs of $k$-ary trees with varying depth $d$ and branching factor $k$; this is akin to generated DAGs using gridded random search (Section~\ref{ssec:gridded_random}) on pipelines of length $d$ for which there are $k$ choices of hyperparameter values at each node.  
We use Gurobi \cite{gurobi}, called via the cvxpy package, to solve the cache-dependent generalized paging problem for each synthetic DAG.

The empirical runtimes of solving for  the optimal caching strategy for different trees are shown in Table~\ref{tab:slowtrees}.  While by no means exhaustive, this table provides guidance for when solving the ILP may be worthwhile. 
The results show that in under 5 minutes one can effectively compute an optimal cache policy for 10s of pipelines.  

\subsubsection{Comparison of Online Cache Strategies}
\label{ssec:cachecomparison}
Given the result in the previous section, it is clear that solving the ILP is  computationally infeasible for larger DAGs where hundreds of pipelines are considered simultaneously. Hence, we explore
online caching strategies, which must make cache decisions based on prior experience and generally have smaller overhead.  Additionally, the computational cost and
memory requirement for all nodes within a 
DAG may not be known beforehand, e.g., when using sequential hyperparameter tuning strategies, in which case an online strategy is
needed.

We compare the optimal strategy (OPT) from solving the ILP to the  online strategies introduced in Section~\ref{sec:related_hp}.
We evaluate the caching algorithms on synthetic DAGs of $k$-ary trees with branching factor $k=3$ and and depth $d=3$.  
In the following experiments, randomized caching strategies are simulated 100 times to reduce variance.

First we consider a scenario where the memory size of each result is  fixed at 10 and the root node is assigned a computational cost of $100$ while all other nodes have a cost of $1$. Figure~\ref{fig:eventreeload} shows the results of this experiment. LRU performs poorly for this configuration precisely because the root node in the tree uses $100$ units of computation and LRU does not account for this. The other strategies converge to the optimal strategy relatively quickly.  

Next, we allow for variable sizes and costs.  Each node is randomly given a size of either 10 or 50 units with equal probability and a corresponding cost of 1 or 100. Figure~\ref{fig:uneventreeload} shows the competitiveness of each strategy vs. the optimal strategy. 
In this setting, LRU is again the least effective policy while WRECIPROCAL is closest to optimal for most cache sizes.  These results on synthetic DAGs suggest that LRU, the default caching strategy in \texttt{scikit-learn} and \texttt{MLlib}, is poorly suited for machine learning pipelines. 
Our experiments in the next section on real-world tasks reinforce this conclusion.


\begin{figure*}[th!]
\centering
\begin{subfigure}{0.3\textwidth}
\includegraphics[width=1.1\textwidth]{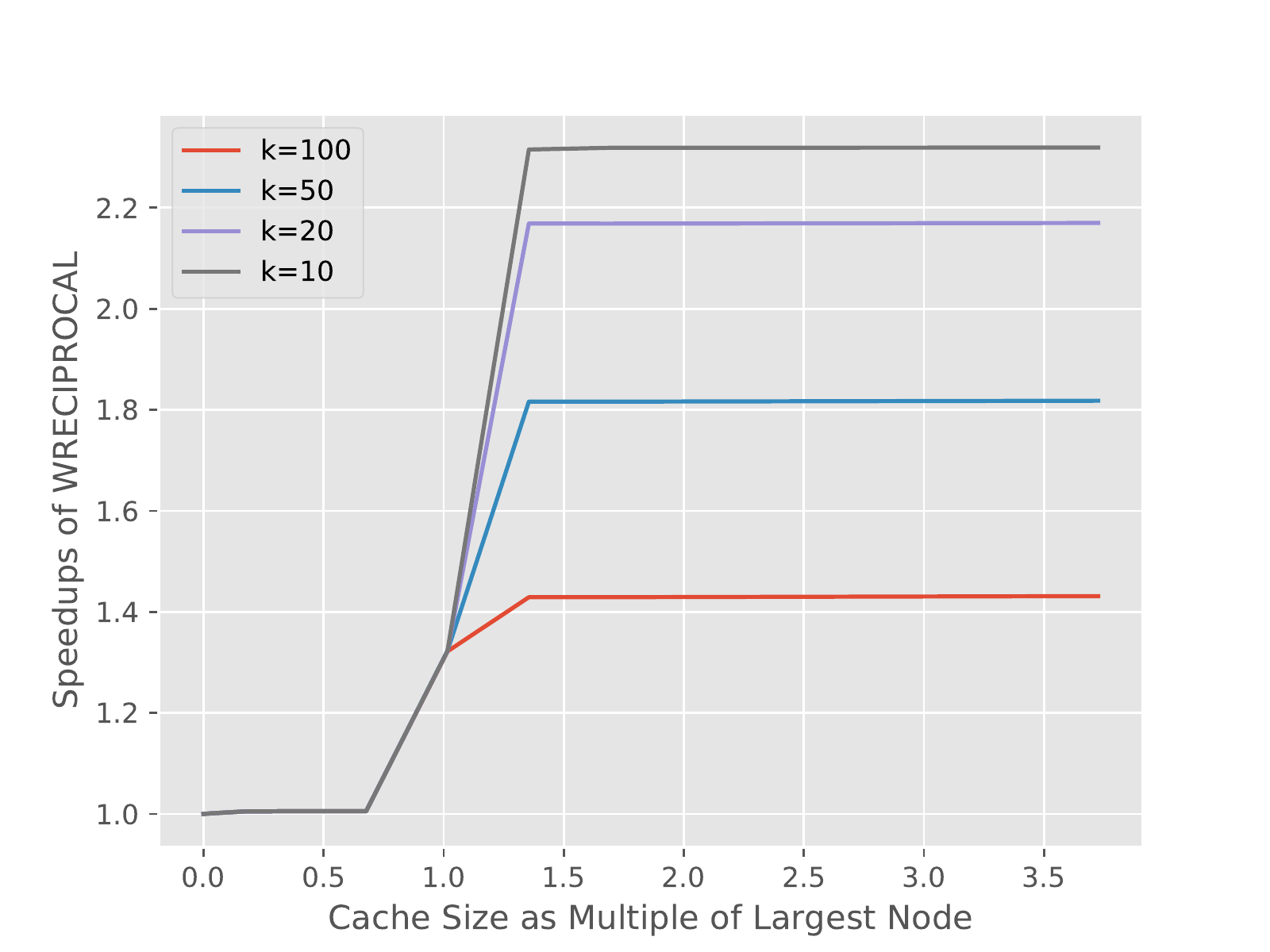}
\end{subfigure}
\begin{subfigure}{0.3\textwidth}
\centering
\includegraphics[width=1.1\textwidth]{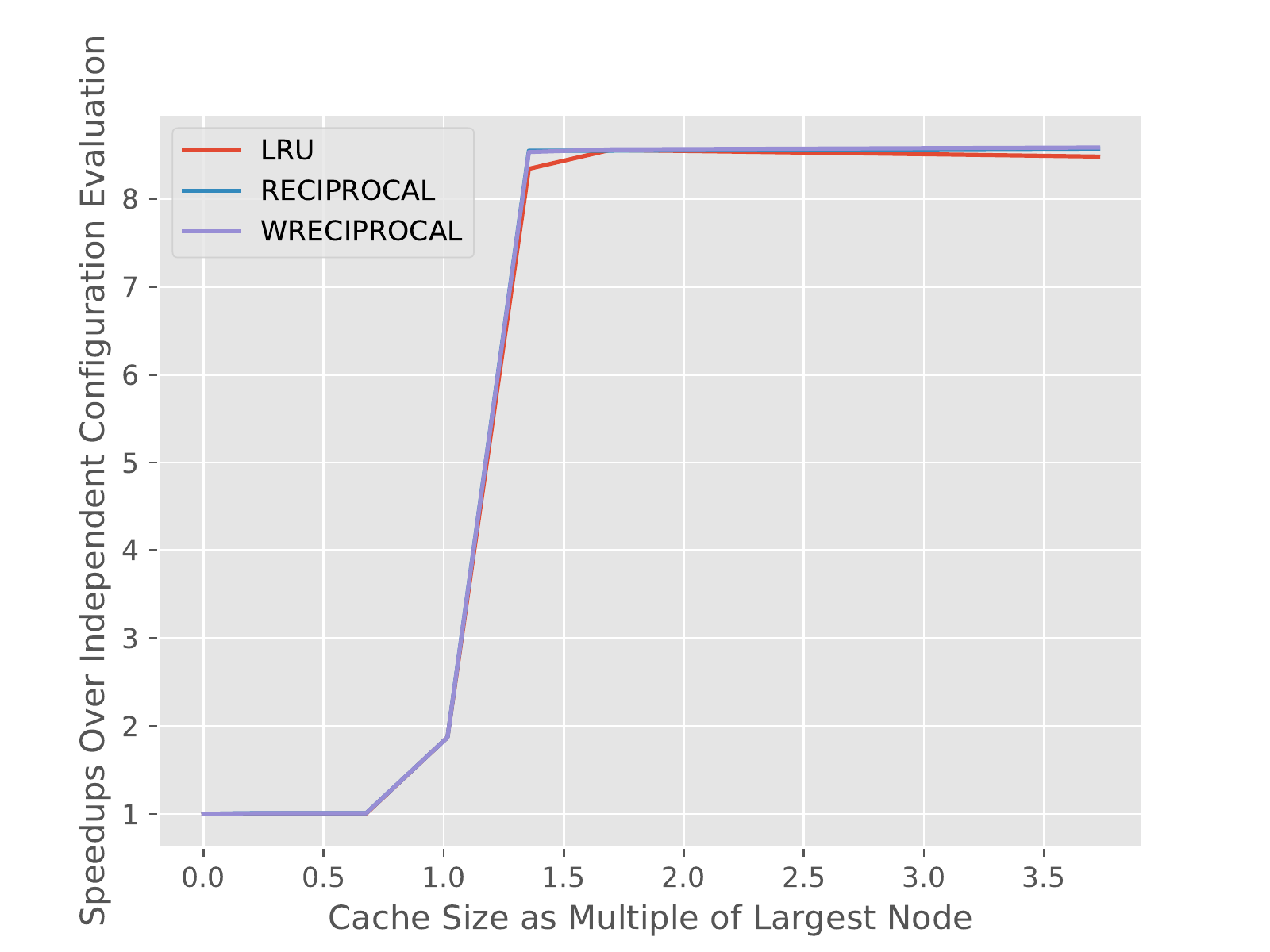}
\end{subfigure}
\begin{subfigure}{0.3\textwidth}
\centering
\includegraphics[width=1.1\textwidth]{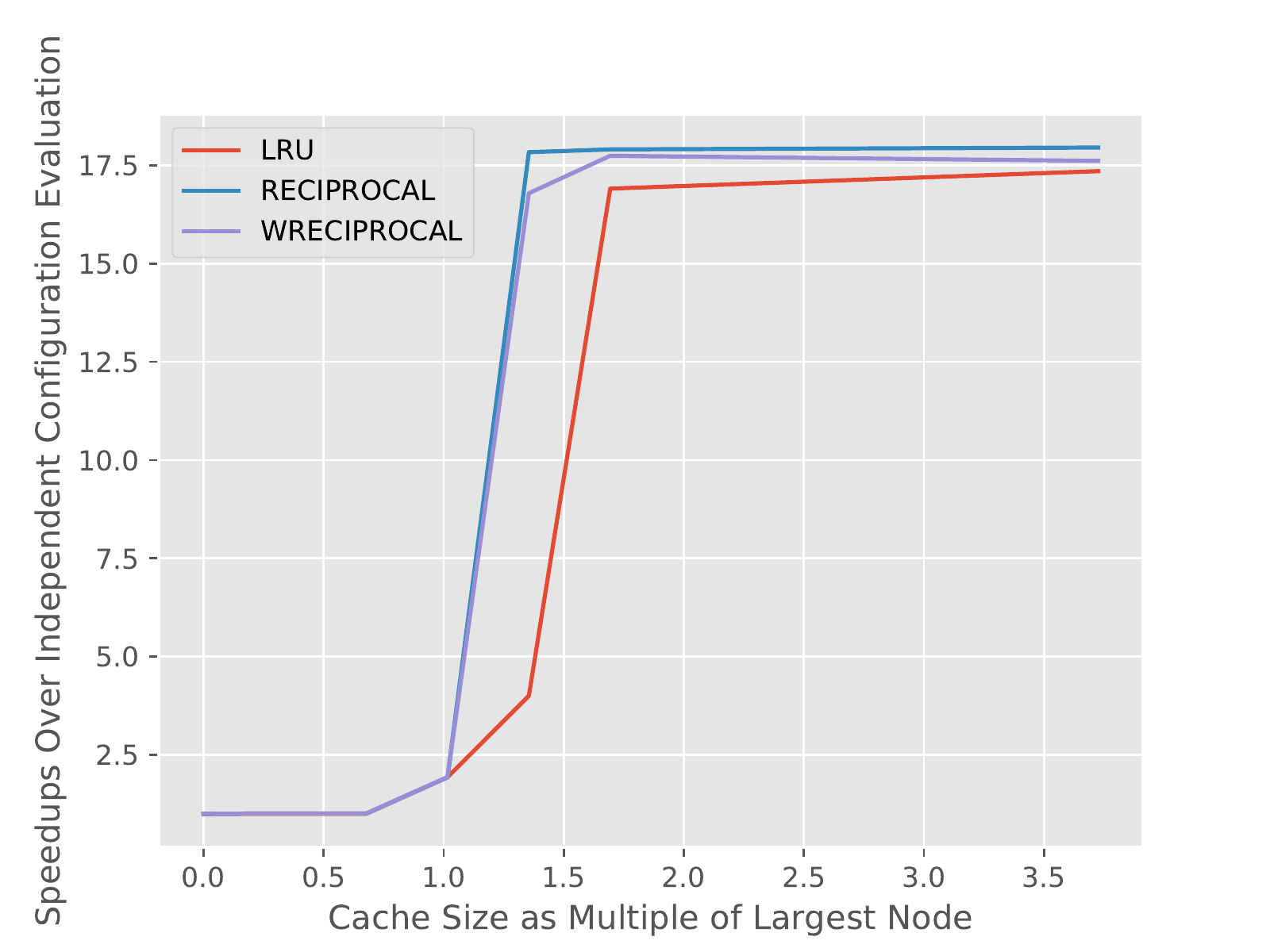}
\end{subfigure}
\caption{Speedups over independent configuration evaluation on TIMIT. 
Left: Speedups using WRECIPROCAL for a DAG with 100 pipelines generated using gridded random search with different branching factors in the random feature step.  As expected, speedups from reuse is higher for DAGs with smaller branching factors. Middle: DAG with 100 pipelines and a maximum downsampling rate of $\nicefrac{R}{64}$ for SH. Right: DAG with 256 pipelines and a maximum downsampling rate of $\nicefrac{R}{256}$ for SH.}
\label{fig:timit}
\end{figure*}

\section{Empirical Studies}
\label{sec:evaluation}
In this section we evaluate the speedups from reuse on real-world hyperparameter tuning tasks gained by integrating our proposed techniques.  
We consider tuning pipelines for the following three datasets: 20 Newsgroups,\footnote{\url{http://qwone.com/~jason/20Newsgroups/}}  Amazon Reviews \citep{amazonreviews}, and the TIMIT~\cite{posen} speech dataset.   Our results are simulated on DAGs generated using profiles collected during actual runs of each pipeline. Details on how the DAGs are created as well as the
search spaces considered for each pipeline are available in Appendix~\ref{appendix:search_space}.  
For each task, we examine potential savings from reuse for 100 pipelines trained simultaneously.  

In Section~\ref{ssec:texttasks}, we examine the speedups from the caching component of pipeline-aware tuning on a text classification pipeline for the Newsgroup and Amazon Reviews datasets.  The DAGs associated with both datasets are naturally amenable to reuse due to discrete hyperparameters and front-loaded computation.  In Section~\ref{ssec:timit}, we then leverage our entire three-pronged approach to pipeline-aware hyperparameter tuning on a speech classification task that at first glance is not amenable to reuse.  

We quantify the speedups by comparing the execution time required for our pipeline-aware approach to the execution time when evaluating the pipelines independently.  We focus on the computational efficiency of our approach as opposed to accuracy, since we have already explored the impact of gridded random search and since SH is known to perform well on a wide range of hyperparameter optimization tasks.  

\subsection{Tuning Text Classification Pipelines}
\label{ssec:texttasks}
We tune the pipeline in Figure~\ref{fig:beforeandafter}(a) for the 20 Newsgroups and the Amazon Reviews datasets.  As shown in Figure~\ref{fig:textpipeline}, 
RECIPROCAL and WRECIPROCAL  offer $70\times$ speedup over independent configuration evaluation for the 20 Newsgroups dataset. Similarly, for the Amazon Reviews dataset, all the caching strategies offer more than $40\times$ speedup over  independent configuration evaluation.  In these instances, caching offers significant speedups due to the front-loaded nature of the pipelines.  By caching, we save on several shared steps: tokenization, NGrams generation, and feature selection/generation.  Additionally, compared to the preprocessing and featurization steps, training is relative inexpensive.  

The frontloaded computation also explains why LRU underperforms on small memory sizes; by biasing older nodes, earlier and more expensive pipeline stages are more likely to be evicted.  Another drawback of LRU is that it does not take into account the cost of computing each node and it will always try to cache the most recent node. Hence, once the size of the cache is large enough to cache the output of the largest node, LRU adds the node to cache and evicts all previous caches, while RECIPROCAL and WRECIPROCAL are less likely to cache the node.  This explains the severe drop in performance for LRU once cache size exceeded 1.  These results also show WRECIPROCAL to be more robust than RECIPROCAL because it is less likely to cache large nodes with relatively inexpensive cost.  Hence, 
we propose using WRECIPROCAL as an alternative to LRU in popular machine learning libraries.

\subsection{Designing DAGs for Reuse on TIMIT}
\label{ssec:timit}
The TIMIT pipeline tuning task performs random feature kernel approximation followed by 20 iterations of LBFGS. The first stage of the pipeline is a continuous hyperparameter, so we first examine the impact of promoting prefix sharing through gridded random search.  Figure~\ref{fig:timit}(left) shows the speedups when using WRECIPROCAL over independent configuration evaluation for gridded random search with different branching factors.  As expected, speedups increase with lower branching factor, though the speedups are muted for all branching factors due to the long training time for this task.  We proceed with a branching factor of 10 for the random feature stage of the pipeline.

Next, we use Successive Halving (Section~\ref{ssec:sha}) with training set size as the resource to evaluate the impact on speedups when reducing the portion of time spent on the last stage of the pipeline. We run SH with the following parameters: $\eta=4$ and $G=4$ (which implies $r=R/64$).  Figure~\ref{fig:timit} (middle) shows the speedups over independent configuration evaluation increases from $2\times$ for uniform allocation to over $8\times$ when using Successive Halving.  Finally, we examine a more realistic setup for SH where the number of pipelines considered is higher to allow for a more aggressive downsampling rate ($n=256$, $G=5$, and $r=R/256$).  Figure~\ref{fig:timit} (right) shows that in this case, speedups from reuse via caching reaches over $17\times$.

\section{Conclusion}
We have provided a thorough consideration of a pipeline-aware approach to hyperparameter tuning and demonstrated that such an approach can offer over an order-of-magnitude speedup on amenable pipelines by exploiting opportunities for reuse.  
Future work includes: (1) formulating a relaxation of the ILP to efficiently approximate the optimal solution for larger pipelines, (2) incorporating meta-learning to predict cost and memory requirements so we can effectively always be in the perfect information setting, and (3) adding alternative cache strategies (e.g., RECIPROCAL and WRECIPROCAL) to popular machine learning libraries like \texttt{scikit-learn} and \texttt{MLlib}.

\bibliography{hypeline}
\bibliographystyle{sysml2019}
\clearpage
\appendix
\section{Appendix}
\label{appendix}
Supplementary material for Section~\ref{sec:promoting_reuse}, Section~\ref{sec:caching} and Section~\ref{sec:evaluation} are shown below.

\subsection{Gridded Random Search Experiments}
\label{appendix:grid}

The datasets were selected from OpenML with the following criteria:
\begin{enumerate}
\item status is active;
\item ratio of majority class to minority class is less than 10;
\item no missing values;
\item not sparse;
\item is a classification task;
\item more than 5k and less than 100k instances;
\item more than 20 and less than 1k features.

\end{enumerate}
We use $2/3$ of the data for training and $1/3$ for testing.  We exclude datasets for which we can learn a perfect classifier on the test set. 

The search space considered for each component is shown in Table~\ref{tab:gridded}.
\begin{table}[h!]
\caption{Hyperparameters considered for the gridded random search experiments in Section~\ref{ssec:gridded_random}.}
\label{tab:gridded}

\centering
\tiny
\begin{tabular}{lccc}
	\\[-1.8ex]\hline
	\hline \\[-1.8ex]

Component & Name & Type & Values \\ \hline
PCA &  variance to keep & continuous & $[0.5, 1.0]$ \\
& whiten & binary & $\{True, False\}$\\ \hline
Select Percentile & percentile to keep & $[0.01, 1.0]$ \\ \hline
FastICA & \# components & discrete & $[10, 2001]$\\ 
& whiten & binary & $\{True, False\}$\\ \hline
Random Forest & min samples to split & discrete & $[2,21]$\\
& min samples per leaf & discrete & $[1, 20]$\\
& bootstrap & binary & $\{True, False\}$\\ \hline
RBF Kernel SVM & kernel scale & continuous & $\log\; [1e-4, 10]$\\
l2 regularizer & continuous & $\log\; [1e-3, 1e3]$
\end{tabular}
\end{table}

\subsection{Details for Experiments in Section~\ref{sec:evaluation}}
\label{appendix:search_space}
We constructed a simulation framework to study the effectiveness of different cache policies at achieving maximal reuse using pipeline-aware hyperparameter tuning.  Our framework estimates the time needed to execute a fully merged hyperparameter pipeline DAG under a given cache policy for different memory sizes. The pipelines enter the simulator as DAGs with all of the necessary metadata required to estimate the execution time for each cache policy.  Specifically, the nodes in the DAG are populated with information about the local  execution time, memory footprint of each node's output, and number of iterations associated with each node using profiled statistics on sample workloads.  These experiments were run using KeystoneML\footnote{\url{keystone-ml.org}} \citep{KeystoneML} with Scala and Apache Spark.  

The search space considered for the 20 Newsgroups pipeline is shown in Figure~\ref{fig:newsgroupsspace}; that for Amazon Reviews is shown in Table~\ref{fig:amazon}; and that for TIMIT is shown in Table~\ref{fig:timitsearchspace}.  We used Apache Spark to profile the computational cost and memory requirement of different pipelines.  We then created DAGs from these profiles to evaluate the different caching strategies.  In order to generate the sub DAGs for Successive Halving, we assumed that training time scales linearly with the dataset size for LBFGS.  

\begin{figure}[t]
\centering
\caption{Search space for Newsgroups workload.}
		\label{fig:newsgroupsspace}
		\tiny
		\begin{tabular}{lll}
	\\[-1.8ex]\hline
	\hline \\[-1.8ex]
		Name      & Type             & Values           \\
		\hline\\
		nGrams    & Integer          & (2,4)            \\
		Top Features  & Integer          & ($10^3$, $10^5$) \\
		Naive Bayes $\lambda$ & Continuous (log) & (0, $10^4$) \\
		\hline
        \\
        \\
		\end{tabular}
		
\centering
\caption{Search space for Amazon Reviews workload.}
		\label{fig:amazon}
				\tiny
		\begin{tabular}{lll}
	\\[-1.8ex]\hline
	\hline \\[-1.8ex]
		Name      & Type             & Values           \\
		\hline\\
		nGrams    & Integer          & (2,4)            \\
		Top Features  & Integer          & ($10^4$, $10^6$) \\
		$\lambda$ & Continuous (log) & ($10^{-5}$, $10^5$) \\
		\hline
        \\
        \\
		\end{tabular}
		
\centering
\caption{Search space for TIMIT workload.}
		\label{fig:timitsearchspace}
				\tiny
		\begin{tabular}{lll}
	\\[-1.8ex]\hline
	\hline \\[-1.8ex]
		Name         & Type             & Values                               \\
		\hline\\
		$\gamma$     & Continuous (log) & ($5.5 \times 10^-4$, $5.5 \times 10^4$) \\
		Distribution & Discrete         & \{Cauchy, Gaussian\}                 \\
		$\lambda$    & Continuous (log) & (0, $10^5$)                         \\
		\hline
		\end{tabular}
		
\end{figure}

\end{document}